\documentclass[11pt,a4paper]{article}
\usepackage[hyperref]{emnlp2019}
\usepackage{times}
\usepackage{latexsym}

\usepackage{graphicx}
\usepackage{wrapfig}
\usepackage{url}
\usepackage{amsmath}
\usepackage{amssymb}
\usepackage{color,xcolor,colortbl}
\usepackage{enumitem}
\usepackage{algorithm}
\usepackage{algorithmic}
\usepackage[font={small}]{caption}
\usepackage{bm,bbm}
\usepackage{booktabs}
\usepackage{mathtools}
\usepackage{array}
\usepackage{multirow}
\usepackage{acronym}
\usepackage{float}

\definecolor{darkblue}{rgb}{0.0, 0.0, 0.55}

\usepackage{soul}

\definecolor{d}{HTML}{c6dbef}
\definecolor{l}{HTML}{e5f5e0}

\aclfinalcopy 

\title{Analyzing Sentence Fusion in Abstractive Summarization}

\author{Logan Lebanoff$^{\spadesuit*}$ \quad John Muchovej$^\spadesuit$\thanks{$^*$These authors contributed equally to this work.} \quad Franck Dernoncourt$^\clubsuit$ \\
\textbf{Doo Soon Kim$^\clubsuit$ \quad Seokhwan Kim$^\heartsuit$ \quad Walter Chang$^\clubsuit$ \quad Fei Liu$^\spadesuit$}\\
\\
$^\spadesuit$University of Central Florida \quad\quad
$^\clubsuit$Adobe Research \quad\quad
$^\heartsuit$Amazon Alexa AI \\[0.3em]
\texttt{\small \{loganlebanoff, john.muchovej\}@knights.ucf.edu \quad feiliu@cs.ucf.edu}\\
\texttt{\small \{dernonco,dkim,wachang\}@adobe.com \quad seokhwk@amazon.com}\\
}

\date{}

\begin{document}
\maketitle

\acrodef{cnn}[CNN/DailyMail]{}
\acrodef{mturk}[MTurk]{Amazon Mechanical Turk}
\acrodef{turkers}[Turkers]{\ac{mturk} evaluators}

\acrodef{pg}[PG]{PG \cite{see_get_2017}}
\acrodef{novel}[Novel]{Novel \cite{kryscinski_improving_2018}}
\acrodef{farl}[Fast-Abs-RL]{Fast-Abs-RL \cite{chen_fast_2018}}
\acrodef{bu}[Bottom-Up]{Bottom-Up \cite{gehrmann_bottom-up_2018}}
\acrodef{dca}[DCA]{DCA \cite{celikyilmaz_deep_2018}}

\begin{abstract}

While recent work in abstractive summarization has resulted in higher scores in automatic metrics, there is little understanding on how these systems combine information taken from multiple document sentences.
In this paper, we analyze the outputs of five state-of-the-art abstractive summarizers, focusing on summary sentences that are formed by sentence fusion. 
We ask assessors to judge the grammaticality, faithfulness, and method of fusion for summary sentences. 
Our analysis reveals that system sentences are mostly grammatical, but often fail to remain faithful to the original article.

\end{abstract}

\section{Introduction}

Modern abstractive summarizers excel at finding and extracting salient content \cite{see_get_2017, chen_fast_2018, celikyilmaz_deep_2018, liu-lapata-2019-hierarchical}. 
However, one of the key tenets of summarization is consolidation of information, and these systems can struggle to combine content from multiple source texts, yielding output summaries that contain poor grammar and even incorrect facts. 
Truthfulness of summaries is a vitally important feature in order for summarization to be widely accepted in real-world applications \cite{Reiter:2018:CL,cao2018faithful}. In this work, we perform an extensive analysis of summary outputs generated by state-of-the-art systems, examining features such as truthfulness to the original document, grammaticality, and method of how sentences are merged together. This work presents the first in-depth human evaluation of multiple diverse summarization models.

We differentiate between two methods of shortening text: sentence {compression} and sentence {fusion}. Sentence compression reduces the length of a \emph{single} sentence by removing words or rephrasing parts of the sentence \cite{cohn-lapata-2008-sentence,  wang-etal-2013-sentence, li-etal-2013-document, li-etal-2014-improving, filippova_sentence_2015}. Sentence fusion reduces \emph{two or more} sentences to one by taking content from each sentence and merging them together \cite{barzilay_sentence_2005, mckeown-etal-2010-time, thadani_supervised_2013}. 
Compression is considered an easier task because unimportant clauses within the sentence can be removed while retaining the grammaticality and truth of the sentence~\cite{mcdonald-2006-discriminative}. 
In contrast, fusion requires selection of important content and stitching of that content in a grammatical and meaningful way. 
We focus on sentence fusion in this work.

\begin{table*}
\setlength{\tabcolsep}{6.6pt}
\renewcommand{\arraystretch}{1.15}
  \begin{center}{\footnotesize
    \begin{tabular}{ l | c c c | c c c c | c }
        \bottomrule[0.3mm]
        \multirow{2}{*}{System} & \multicolumn{3}{c|}{ROUGE} & \multicolumn{4}{c|}{Created By} &  Avg Summ\\
         	            &	R-1	    &	R-2	    &	R-L	    & Compress  &	Fuse	&	Copy	&	Fail	&	 Sent Len\\
         	\hline
            \acl{pg}	&	39.53	&	17.28	&	36.38	&	63.14	&	\,\;6.44	&	\textbf{30.24}	&	\,\;0.18	&	15.7\\
            \acl{novel} &	40.19	&	17.38	&	37.52	&	71.25	&	19.77	&	\,\,5.39	&	\,\,3.59	&	11.8\\
            \acl{farl}	&	40.88	&	17.80	&	\textbf{38.54}	&	\textbf{96.65}	&	\,\,0.83	&	\;\;2.21	&	\,\;0.31	&	15.6\\
            \acl{bu}	&	41.22	&	18.68	&	38.34	&	71.15	&	16.35	&	11.76	&	\,\;0.74	&	10.7\\
            \acl{dca}	&	\textbf{41.69}	&	\textbf{19.47}	&	37.92	&	64.11	&	23.96	&	\,\;7.07	&	\,\;4.86	&	14.5\\
            Reference Summaries	&	-	    &	-	    &	-	    &	60.65	&	\textbf{31.93}	&	\,\;1.36	&	\,\;\textbf{6.06}	&	\textbf{19.3}\\
        \toprule[0.3mm]
      \end{tabular}}
  \end{center}
  \vspace{-0.1in}
  \caption{Comparison of state-of-the-art summarization systems. Middle column describes how summary sentences are generated. \textit{Compress}: single sentence is shortened. \textit{Fuse}: multiple sentences are merged. \textit{Copy}: sentence is copied word-for-word. \textit{Fail}: did not find matching source sentences.}
  \vspace{-0.15in}
  \label{tab:characteristics} 
\end{table*}

We examine the outputs of five abstractive summarization systems on \acs{cnn} \cite{hermann_teaching_2015} using human judgments. 
Particularly, we focus on summary sentences that involve sentence fusion, since fusion is the task that requires the most improvement. We analyze several dimensions of the outputs, including faithfulness to the original article, grammaticality, and method of fusion. We present three main findings:

\begin{itemize}[topsep=3pt,itemsep=-1pt,leftmargin=*]
\item 38.3\% of the system outputs introduce incorrect facts, while 21.6\% are ungrammatical;

\item systems often simply concatenate chunks of text when performing sentence fusion, while largely avoiding other methods of fusion like entity replacement;

\item systems struggle to reliably perform complex fusion, as entity replacement and other methods result in incorrect facts 47--75\% of the time.
\end{itemize}


\section{Evaluation Setup}
\label{sec:human-analysis}

Evaluation of summarization systems relies heavily on automatic metrics. 
However, ROUGE~\cite{lin-2004-rouge} and other n-gram based metrics are limited in evaluation power and do not tell the whole story \cite{novikova-etal-2017-need}. 
They often focus on informativeness, which misses out on important facets of summaries such as faithfulness and grammaticality. 
In this paper we present a thorough investigation of several abstractive summarization systems using human evaluation on CNN/DailyMail. 
The task was accomplished via the crowdsourcing platform 
\acl{mturk}.
We particularly focus on summary sentences formed by sentence fusion, as it is arguably a harder task and is a vital aspect of abstractive summarization.

\subsection{Summarization Systems}

We narrowed our evaluation to five state-of-the-art summarization models\footnote{The summary outputs from PG, Bottom-Up, and Fast-Abs-RL are obtained from their corresponding Github repos. Those from Novel and DCA are graciously provided to us by the authors. We thank the authors for sharing their work.}, as they represent some of the most competitive abstractive summarizers developed in recent years.
The models show diversity across several dimensions, including ROUGE scores, abstractiveness, and training paradigm. We briefly describe each system, along with a comparison in Table \ref{tab:characteristics}.

\begin{itemize}[topsep=5pt,itemsep=0pt,leftmargin=*]
\item \textbf{PG}~\cite{see_get_2017} The pointer-generator networks use an encoder-decoder architecture with attention and copy mechanisms that allow it to either generate a new word from the vocabulary or copy a word directly from the document. It tends strongly towards extraction and copies entire summary sentences about 30\% of the time.

\item \textbf{Novel}~\cite{kryscinski_improving_2018} This model uses an encoder-decoder architecture but adds a novelty metric which is optimized using reinforcement learning. 
It improves summary novelty by promoting the use of unseen words.

\item \textbf{Fast-Abs-RL}~\cite{chen_fast_2018} Document sentences are selected using reinforcement learning and then compressed/paraphrased using an encoder-decoder model to generate summary sentences. 

\item \textbf{Bottom-Up}~\cite{gehrmann_bottom-up_2018} An external content selection model identifies which words from the document should be copied to the summary; such info is incorporated into the copy mechanism of an encoder-decoder model.

\item \textbf{DCA}~\cite{celikyilmaz_deep_2018} The source text is divided among several encoders, which are all connected to a single decoder using hierarchical attention. It achieves one of the highest ROUGE scores among state-of-the-art.

\end{itemize}

\begin{figure*}
    \centering
    \includegraphics[width=0.99\textwidth]{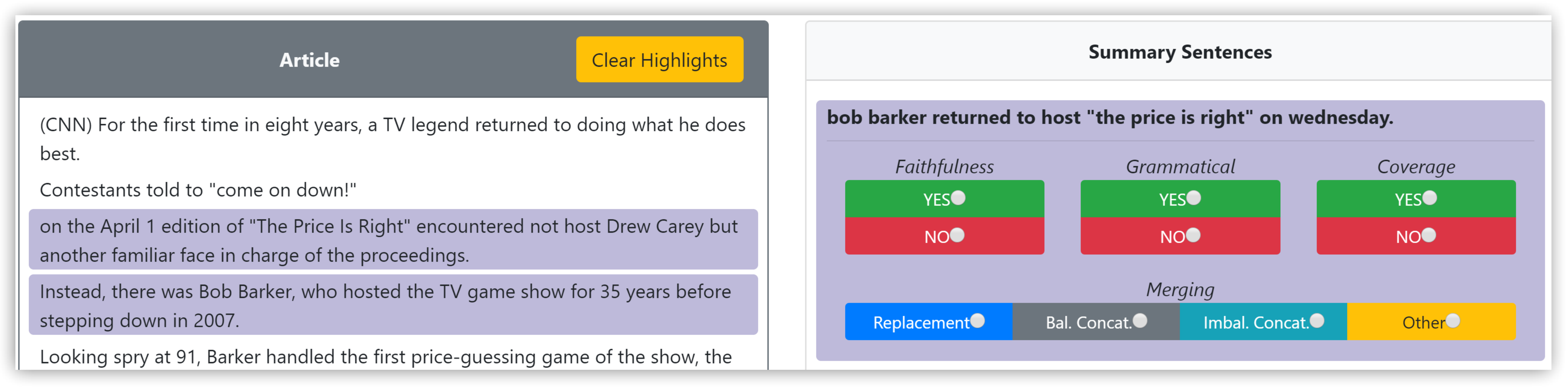}
    \caption{Annotation interface. A sentence from a random summarization system is shown along with four questions.}
    \label{fig:interface}
    \vspace{-0.05in}
\end{figure*}

\subsection{Task Design}
\label{sec:human-analysis-task-design}

Our goal is to assess the quality of summary sentences according to their grammaticality, faithfulness and method of fusion.
We design a crowd task consisting of a single article with six summary sentences:
one sentence is guaranteed to be from the reference summary, the other five are taken from system summaries.
An annotator is instructed to read the article, then rate the following characteristics for each summary sentence:

\paragraph{Faithfulness}
For a summary to be useful, it must remain true to the original text. This is particularly challenging for abstractive systems since they require a deep understanding of the document in order to rephrase sentences with the same meaning.

\paragraph{Grammaticality}
System summaries should follow grammatical rules in order to read well. Maintaining grammaticality can be relatively straightforward for sentence compression, as systems generally succeed at removing unnecessary clauses and interjections \cite{see_get_2017}. However, sentence fusion requires greater understanding in order to stitch together clauses in a grammatical way.

\paragraph{Method of Merging}
Each summary sentence in our experiments is created by fusing content from two document sentences. We would like to understand how this fusion is performed. The following possibilities are given:
\begin{itemize}[topsep=5pt,itemsep=0pt,leftmargin=*]
    \item \textit{Replacement:} a pronoun or description of an entity in
    one sentence is replaced by a different description of that
    entity in the other sentence.
    \item \textit{Balanced concatenation:} a consecutive part of one
    sentence is concatenated with
    a consecutive part of the other sentence. The parts taken
    from each sentence are of similar length.
    \item \textit{Imbalanced concatenation:} similar to the case of ``balanced
    concatenation,'' but the part taken from one sentence is
    larger than the part taken from the other sentence.
    \item \textit{Other:} all remaining cases.
\end{itemize}

\vspace{0.1in}
\paragraph{Coverage}
An annotator is asked to rate how well highlighted article sentences ``covered'' the information contained in the summary sentence. 
Two article sentences that best match a summary sentence are selected according to a heuristic developed by \citet{lebanoff2019scoring}.
The same heuristic is also used to determine whether a summary sentence is created by compression or fusion (more details later in this section). 
Given the importance of this heuristic for our task, we would like to measure its effectiveness on selecting article sentences that best match a given summary sentence.

\vspace{0.1in}
We provide detailed instructions, including examples and explanations. 
We randomly select 100 articles from the CNN/DailyMail test set. This results in 100 tasks for annotators, where each task includes an article and six summary sentences to be evaluated---one of which originates from the reference summary and the other five are from any of the system summaries.
Each task is completed by an average of 4 workers. All workers are required to have the ``Master'' qualification, a designation for high-quality annotations.
Of the 600 summary sentences evaluated, each state-of-the-art system
contributes as follows---\emph{\acs{bu}}: 146, \emph{\acs{dca}}: 130, \emph{\acs{pg}}: 37,
\emph{\acs{novel}}: 171, \emph{\acs{farl}}: 16, and \emph{Reference}: 100.
The number of sentences we evaluate for each system is proportional to the number of observed fusion cases.

In order to answer the \textit{Method of Merging} and \textit{Coverage} questions, the annotator must be provided with which two article sentences were fused together to create the summary sentence in question. We use the heuristic proposed by \citet{lebanoff2019scoring} to estimate which pair of sentences should be chosen. 
They use averaged ROUGE-1, -2, -L scores~\cite{lin-2004-rouge} to represent sentence similarity.
The heuristic calculates the ROUGE similarity between the summary sentence and each article sentence. The article sentence with the highest similarity is chosen as the first sentence, then overlapping words are removed from the summary sentence. It continues to find the article sentence most similar to the remaining summary sentence, which is chosen as the second sentence. Our interface automatically highlights this pair of sentences (Figure \ref{fig:interface}). 

The same heuristic is also employed in deciding whether a summary sentence was generated by sentence compression or fusion. The algorithm halts if no article sentence is found that shares two or more content words with the summary sentence. If it halts after only one sentence is found, then it is classified as \textit{compression}. If it finds a second sentence, then it is classified as \textit{fusion}.

\section{Results}
\label{sec:human-analysis-results}

\begin{table}[t]
\setlength{\tabcolsep}{6.5pt}
\renewcommand{\arraystretch}{1.15}
  \begin{center}{\footnotesize
    \begin{tabular}{ l | c c c }
        \bottomrule[0.3mm]
        \textbf{System}      & \textbf{Faithful} & \textbf{Grammatical} &\textbf{Coverage} \\
        \hline
        \acs{dca}	& 47.0	& 72.4	& 62.6 \\
        \acs{bu}	& 56.9	& 78.9	& 78.5 \\
        \acs{novel}	& 58.5	& 78.5	& 75.3 \\
        \acs{farl}	& 69.0	& 77.6	& 82.8 \\
        \acs{pg}    & {76.9}	& {84.6}	& {89.5} \\
        \hline
        Reference	& 88.4	& 91.6	& 74.9 \\
        \toprule[0.3mm]
      \end{tabular}}
  \end{center}
  \vspace{-0.1in}
  \caption{Percentage of summary sentences that are faithful, grammatical, etc. according to human evaluation of several state-of-the-art summarization systems (see \S\ref{sec:human-analysis} for details). }
  \label{tab:averages} 
  \vspace{-0.1in}
\end{table}

We present experimental results in Table \ref{tab:averages}. 
Our findings suggest that system summary sentences formed by fusion have low faithfulness (61.7\% on average) as compared to the reference summaries. 
This demonstrates the need for current summarization models to put more emphasis on improving the faithfulness of generated summaries. 
Surprisingly, the highest performing systems, DCA and Bottom-Up, according to ROUGE result in the lowest scores for being faithful to the article. 
While we cannot attribute the drop in faithfulness to an over-emphasis on optimizing automatic metrics, we can state that higher ROUGE scores does not necessarily lead to more faithful summaries, as other works have shown \cite{falke_ranking_2019}. 
Bottom-Up, interestingly, is 20 points lower than PG, which it is closely based on. It uses an external content selector to choose what words to copy from the article. While identifying summary-worthy content improved ROUGE, we believe that Bottom-Up stitches together sections of content that do not necessarily belong together. Thus, it is important to identify not just summary-worthy content, but also \textit{mergeable} content.

System summary sentences created by fusion are generally grammatical (78.4\% on average), though it is still not up to par with reference summaries (91.6\%). The chosen state-of-the-art systems use the encoder-decoder architecture, which employs a neural language model as the decoder, and language models generally succeed at encoding grammar rules and staying fluent \cite{clark-etal-2019-bert}. The coverage for reference summaries is moderately high (74.9\%), demonstrating the effectiveness of the heuristic of identifying where summary content is pulled from. Especially for most of the systems, the heuristic successfully finds the correct source sentences. As it is based mostly on word overlap, the heuristic works better on summaries that are more extractive, hence the higher coverage scores among the systems compared to reference summaries, which are more abstractive.

\begin{figure}
    \centering
    \includegraphics[width=0.95\linewidth]{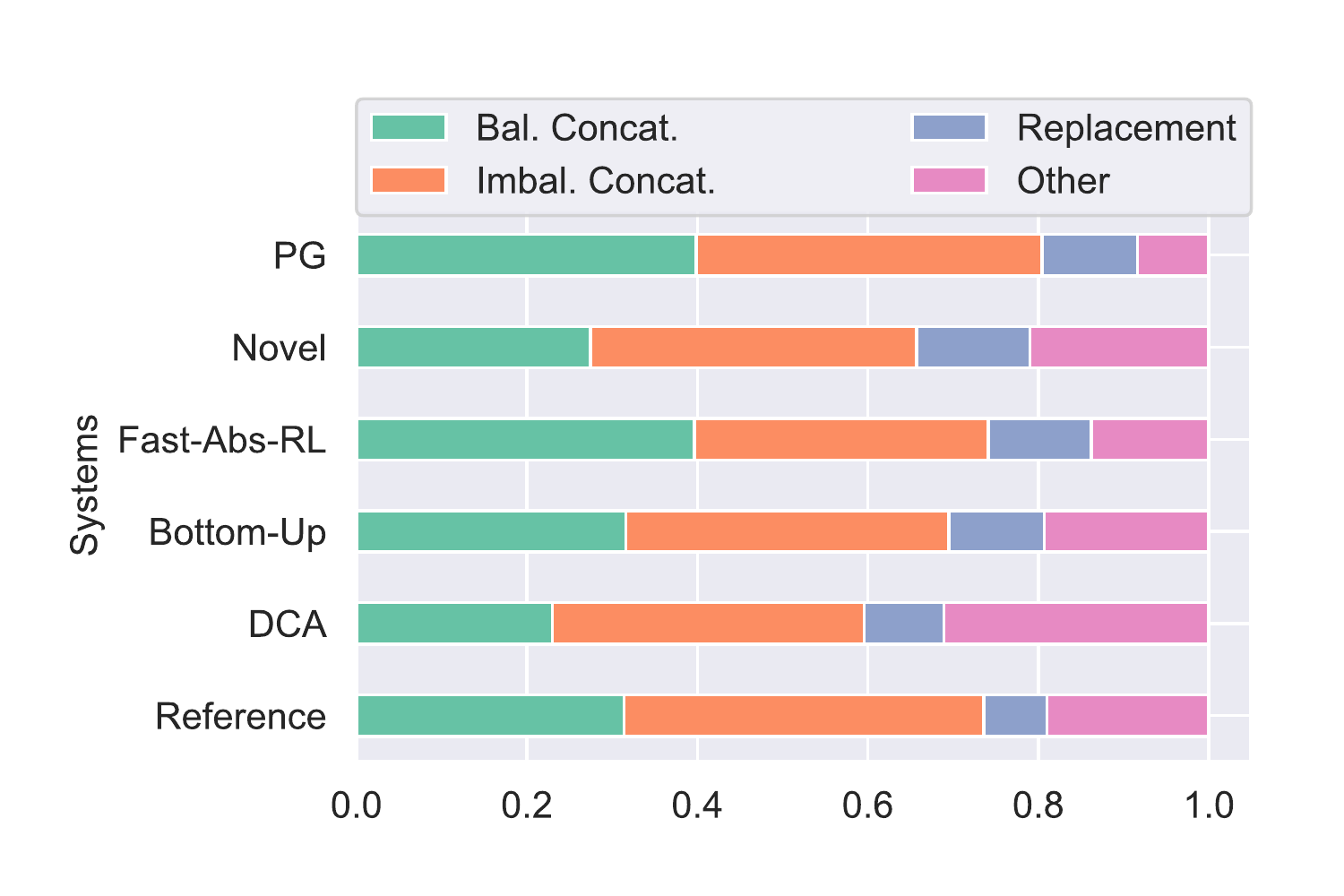}
    \caption{Frequency of each merging method. Concatenation is the most common method of merging.}
    \label{fig:merge_dist}
\end{figure}

Figure \ref{fig:merge_dist} illustrates the frequency of each merging method over the summarization systems. Most summary sentences are formed by concatenation. PG in particular most often fuses two sentences using concatenation. Surprisingly, very few reference summaries use entity replacement when performing fusion. We believe this is due to the extractiveness of the CNN/DailyMail dataset, and would likely have higher occurrences in more abstractive datasets.

Does the way sentences are fused affect their faithfulness and grammaticality? Table \ref{tab:merge-averages} provides insights regarding this question. Grammaticality is relatively high for all merging categories. Coverage is also high for balanced/imbalanced concatenation and replacement, meaning the heuristic works succesfully for these forms of sentence merging. It does not perform as well on the Other category. This is understandable, since sentences formed in a more complex manner will be harder to identify using simple word overlap. Faithfulness has a similar trend, with summaries generated using concatenation being more likely to be faithful to the original article. This may explain why PG is the most faithful of the systems, while being the simplest---it uses concatenation more than any of the other systems. We believe more effort can be directed towards improving the more complex merging paradigms, such as entity replacement.

\begin{table}
\setlength{\tabcolsep}{5.5pt}
\renewcommand{\arraystretch}{1.15}
  \begin{center}{\footnotesize
    \begin{tabular}{ l | c c c }
        \bottomrule[0.3mm]
        \textbf{System}      & \textbf{Faithful} & \textbf{Grammatical} &\textbf{Coverage} \\
        \hline
        Bal Concat	& 82.55	& 86.91	& 94.43  \\
        Imbal Concat	& 69.40	& 80.25	& 84.58 \\
        Replacement	& 53.06	& 82.04	& 77.55 \\
        Other	        & 25.20	& 68.23	& 27.04 \\
        \toprule[0.3mm]
      \end{tabular}}
  \end{center}
  \vspace{-0.1in}
  \caption{Results for each merging method. Concatenation has high faithfulness, grammaticality, and coverage, while Replacement and Other have much lower scores.}
  \label{tab:merge-averages} 
  \vspace{-0.05in}
\end{table}

There are a few potential limitations associated with the experimental design.
Judging whether a sentence is faithful to the original article can be a difficult task to perform reliably, even for humans. 
We observe that the reference summaries achieve lower than the expected faithfulness and grammaticality of 100\%. 
This can have two reasons.
First, the inter-annotator agreement for this task is relatively low and we counteract this by employing an average of four annotators to complete each task.
Second, we make use of an automatic heuristic to highlight sentence pairs from the article. While it generally finds the correct sentences---average Coverage score of 77.3\%---the incorrect pairs may have biased the annotators away from sentences that humans would have found more appropriate.
This further exemplifies the difficulty of the task.

\section{Related Work}

Sentence fusion aims to produce a single summary sentence by fusing multiple source sentences.
Dependency graphs and discourse structure have proven useful for aligning and combining multiple sentences into a single sentence \cite{barzilay_sentence_2005, marsi_explorations_2005, filippova_sentence_2008, cheung_unsupervised_2014, gerani-etal-2014-abstractive}. 
\citet{mehdad_abstractive_2013} construct an entailment graph over sentences for sentence selection, then fuse sentences together using a word graph. Abstract meaning representation and other graph-based representations have also shown success in sentence fusion \cite{liu-etal-2015-toward, nayeem_abstractive_2018}. \citet{geva_discofuse:_2019} fuse pairs of sentences together using Transformer, focusing on discourse connectives between sentences.

Recent summarization research has put special emphasis on faithfulness to the original text. \citet{cao_retrieve_2018} use seq-to-seq models to rewrite templates that are prone to including irrelevant entities. Incorporating additional information into a seq-to-seq model, such as entailment and dependency structure, has proven successful \cite{li_ensure_2018, song_structure-infused_2018}. The closest work to our human evaluation seems to be from \citet{falke_ranking_2019}. Similar to our work, they find that the PG model is more faithful than Fast-Abs-RL and Bottom-Up, even though it has lower ROUGE. They show that 25\% of outputs from these state-of-the-art summarization models are unfaithful to the original article. \citet{cao2018faithful} reveal a similar finding that 27\% of the summaries generated by a neural sequence-to-sequence model have errors. Our study, by contrast, finds 38\% to be unfaithful, but we limit our study to only summary sentences created by \emph{fusion}. 
Our work examines a wide variety of state-of-the-art summarization systems, and perform in-depth analysis over other measures including grammaticality, coverage, and method of merging.

\section{Conclusion}

In this paper we present an investigation into sentence fusion for abstractive summarization. Several state-of-the-art systems are evaluated, and we find that many of the summary outputs generate false information. Most of the false outputs were generated by entity replacement and other complex merging methods. These results demonstrate the need for more attention to be focused on improving sentence fusion and entity replacement.

\section*{Acknowledgments}

We are grateful to the anonymous reviewers for their helpful comments and suggestions.
This research was supported in part by the National Science Foundation grant IIS-1909603.

\bibliography{logan_summ,fei_summ}
\bibliographystyle{acl_natbib}

\end{document}